\pdfoutput=1

\documentclass[11pt]{article}

\usepackage[]{acl}

\usepackage{times}
\usepackage{latexsym}
\usepackage{graphicx}
\usepackage[T1]{fontenc}

\usepackage[utf8]{inputenc}

\title{HausaNLP at SemEval-2023 Task 12: Leveraging African Low Resource TweetData for Sentiment Analysis}

\author{Saheed Abdullahi Salahudeen$^{1}$, {\bf Falalu Ibrahim Lawan$^{1}$}, Ahmad Mustapha Wali$^{2}$,\\ 
\bf Amina Abubakar Imam$^{3}$, Aliyu Rabiu Shuaibu$^{4}$, Aliyu Yusuf$^{5}$, Nur Bala Rabiu$^{6}$, \\
\bf Musa Bello$^{1}$, Shamsuddeen Umaru Adamu$^{1}$, Saminu Mohammad Aliyu$^{6}$,\\
{\bf  Murja Sani Gadanya$^{6}$, Sanah Abdullahi Muaz$^{6}$}, {\bf Mahmoud Said Ahmad}, \\
  {\bf Abdulkadir Abdullahi$^{7}$}, {\bf Abdulmalik Yusuf Jamoh$^{8}$} \\[2mm]
  $^\forall$HausaNLP, $^{1}$Kaduna State University, $^{2}$University of Bucharest, $^{3}$University of Abuja, \\ $^{4}$Nile University, $^{5}$Universiti Teknologi PETRONAS, $^{6}$Bayero University Kano, \\
  $^{7}$Shehu Shagari College of Education, $^{8}$Ahmadu Bello University, Zaria \\[2mm]
  \texttt{a.salahudeen@kasu.edu.ng} \\}

\begin{document}
  
\maketitle

\begin{abstract}
We present the findings of SemEval-2023 Task 12, a shared task on sentiment analysis for low-resource African languages using Twitter dataset. The task featured three subtasks; subtask A is monolingual sentiment classification with 12 tracks which are all monolingual languages, subtask B is multilingual sentiment classification using the tracks in subtask A and subtask C is a zero-shot sentiment classification. We present the results and findings of subtask A, subtask B and subtask C. We also release the code on github. Our goal is to leverage low-resource tweet data using pre-trained Afro-xlmr-large, AfriBERTa-Large, Bert-base-arabic-camelbert-da-sentiment (Arabic-camelbert), Multilingual-BERT (mBERT) and BERT models for sentiment analysis of 14 African languages. The datasets for these subtasks consists of a gold standard multi-class labeled Twitter datasets from these languages. Our results demonstrate that Afro-xlmr-large model performed better compared to the other models in most of the languages datasets. Similarly, Nigerian languages: Hausa, Igbo, and Yoruba achieved better performance compared to other languages and this can be attributed to the higher volume of data present in the languages.
\end{abstract}

\section{Introduction}
Social media offers an opinionated platform of data content on many topics of interest, such as product reviews, feedback on purchases or services, political interest, etc., that is dynamically created by users in different languages. Sentiment analysis in many languages is required due to the need for effective classification of these contents. However, the majority of research on sentiment analysis has been done in high-resource languages \cite{r1_jha2016generating,r2_vilares2016cs,r3_heikal2018sentiment,r4_elhadad2019sentiment,abdullahi2021deep,r6_carvalho2021evaluation} while several low-resource African languages receive little attention in Natural Language Processing (NLP) application due to insufficient of publicly available data. Although, recently, some considerable efforts are made for the development of sentiment analysis for some low-resource African languages \cite{r7_devlin2018bert,r8_yimam2020exploring,r9_ogueji2021small,r10_abubakar2021enhanced,muhammad-etal-2022-naijasenti}, nonetheless, sentiment analysis for low-resourced African languages still is a misrepresented research area.

\par The AfriSenti\footnote{\url{https://github.com/afrisenti-semeval}} shared task 12 \cite{muhammadSemEval2023} aims at building sentiment analysis for 14 low-resource African languages using a Twitter dataset which include Hausa, Yoruba, Igbo, Nigerian Pidgin from Nigeria, Amharic, Tigrinya, and Oromo from Ethiopia, Swahili from Kenya and Tanzania, Algerian Arabic dialect from Algeria, Kinyarwanda from Rwanda, Twi from Ghana, Mozambique Portuguese from Mozambique  and Moroccan Arabic/Darija from Morocco. The task featured the following subtasks: Subtask A (Monolingual Sentiment Classification): Given a training document, analyze the sentiment classification of 12 individual African languages. Subtask B (Multilingual Sentiment Classification): Given a training document, analyze the sentiment classification of multiple languages using the languages in task A. Subtask C (Zero-Shot Sentiment Classification): Given evaluation data only, analyze the sentiment classification of only two African languages.

\par In this paper, we demonstrate our approach to tackling the concerns raised in SemEval Task 12 as well as the obstacles posed by low-resource languages \cite{muhammad2023afrisenti}. We used the Bert, AfriBERTa\_large and Afro-xlmr-large model to facilitate the classification of tweets using a monolingual, multilingual and zero shot approach. The rest of the paper is organized as follows. Section 2 is the related works. Section 3 describes the proposed approach. Experimentation and evaluation are discussed in section 4, while section 5 draws some conclusions and discusses some directions for future work.

\section{Related Work}
Sentiment Analysis is the process of generating positive or negative sentiment from data using computational techniques, and to some extent is capable of predicting and classifying tempers such as excitement, anger, and sadness\cite{r14_fortin2019multimodal}. Sentiment and emotion are considered to be related, \cite{r15_akhtar2019multi}therefore, the process of assigning polarity to text and emotion among others has become a prevalent task in NLP. \cite{r16_moore2020multi}. Hence, sentiment analysis is to generate opinions based on a given input provided by users \cite{r17_mccann2018natural}, the current campaign in NLP is that sentiment analysis is extensively adopted for opinion mining to collect information about users from different fields of a particular aspect.

\cite{r10_abubakar2021enhanced} authors use a machine learning approach to combine English and Hausa features to measure classification performance and create a more precise sentiment classification process. \cite{r8_yimam2020exploring,r9_ogueji2021small}  demonstrated that training is feasible with less than 1GB of text to build a competitive multilingual language model. Their results show that "smalldata" technique which uses languages that are similar to one another may occasionally be more effective than combined training on big datasets with high-resource languages.

Although sentiment analysis has been extensively used in many high-resource languages like English and French just to mention a few, little attention is paid to African low-resource languages. \cite{muhammad-etal-2022-naijasenti} presented the first extensive human-annotated Twitter sentiment dataset for the Hausa, Igbo, Nigerian-Pidgin, and Yoruba languages—the four most widely spoken in Nigeria—consisting of roughly 30,000 annotated tweets per language (and 14,000 for Nigerian-Pidgin) and a sizable portion of code-mixed tweets. For these low-resource languages, they suggested text collecting, filtering, processing, and labeling techniques that let us build datasets. They used a variety of pre-trained models and transfer methods. They discovered that the most effective methods are language-specific models and language-adaptive fine-tuning.

\section{Shared Task Description}
The AfriSenti-SemEval Shared Task 12 is based on a collection of Twitter datasets in 14 African languages for sentiment classification. Participants are provided with a training dataset and are required to make a prediction using multiclass sentiment classification. It consists of three sub-tasks: Monolingual sentiment classification, multilingual sentiment classification, and zero-shot sentiment classification. In this paper, we concentrate on all the three subtasks with a total of 14 languages.

\subsection{Subtask A: Monolingual Sentiment Classification}
For this subtask, we used a single language for sentiment classification. Given training data in a target language, we determine the polarity of a tweet in the target language (positive, negative, or neutral). If a tweet conveys both a positive and negative sentiment, whichever is the stronger sentiment should be chosen. This subtask consists of 12 African languages: Hausa, Yoruba, Igbo, Nigerian-Pidgin, Amharic, Algerian Arabic, Moroccan Arabic/Darija, Swahili, Kinyarwanda, Twi, Mozambican Portuguese, Xitsonga (Mozambique dialect).  For this subtask, the dataset is split into 70\% training and 30\% validation. We select an individual language, fine-tune the models with the provided training data, and fine-tune several hyper-parameters to obtain the optimal performances using only 2 models: Afro-xlmr-large and Bert-base-arabic-camelbert-da-sentiment. Afro-xlmr-large is used in all the languages with the exception of  Darija and Algerian Arabic and this is because Afro-xlmr-large model was not trained using the two languages, therefore, we used the Bert-base-arabic-camelbert-da-sentiment for these 2 languages.

\subsection{Subtask B: Multilingual Sentiment Classification}
Given combined training data from Subtask-A (Track 1 to 12), we determine the polarity of a tweet in the target language (positive, negative, or neutral). For this subtask, the multilingual dataset is split into three parts: 90\% training, 5\% each for validation and test set. The three-part split allows for a more robust evaluation of the model, avoids overfitting, and produces a better estimate of the model's generalization performance. We implemented only 1 model for this subtask: Afroxlmr-large-ner-masakhaner-1.0-2.0 and this is because the model was trained using almost all the African languages.

\subsection{Subtask C: Zero Shot Classification}
Zero-shot learning is a type of machine learning that allows a model to perform a task on a new data
point without having been trained on any data points from that class. As given in Subtask C, working
with languages that have limited resources. Zero-shot learning can be used to classify the sentiment
of text in two non-labelled African languages, Tigrinya and Oromo. The implementation of zero-shot
learning for sentiment analysis is to use a multilingual language model like AfroXLM-R, which has
been pre-trained on a large corpus of text from multiple African languages. The pre-trained language
model, would learn to understand the underlying patterns in language across the different languages
and use this knowledge to classify text in the 2 new languages.

However, limitations such as the reliance on language similarity and the assumption that the learned
representations can transferable across languages might cause it to underperform due to the fact
that the target language can be significantly different from the languages in the pre-training corpus
\cite{wang2021learning}

\subsection{Dataset Description}
\begin{figure*}[!t]
    \centering
   \includegraphics[width=0.7\linewidth]{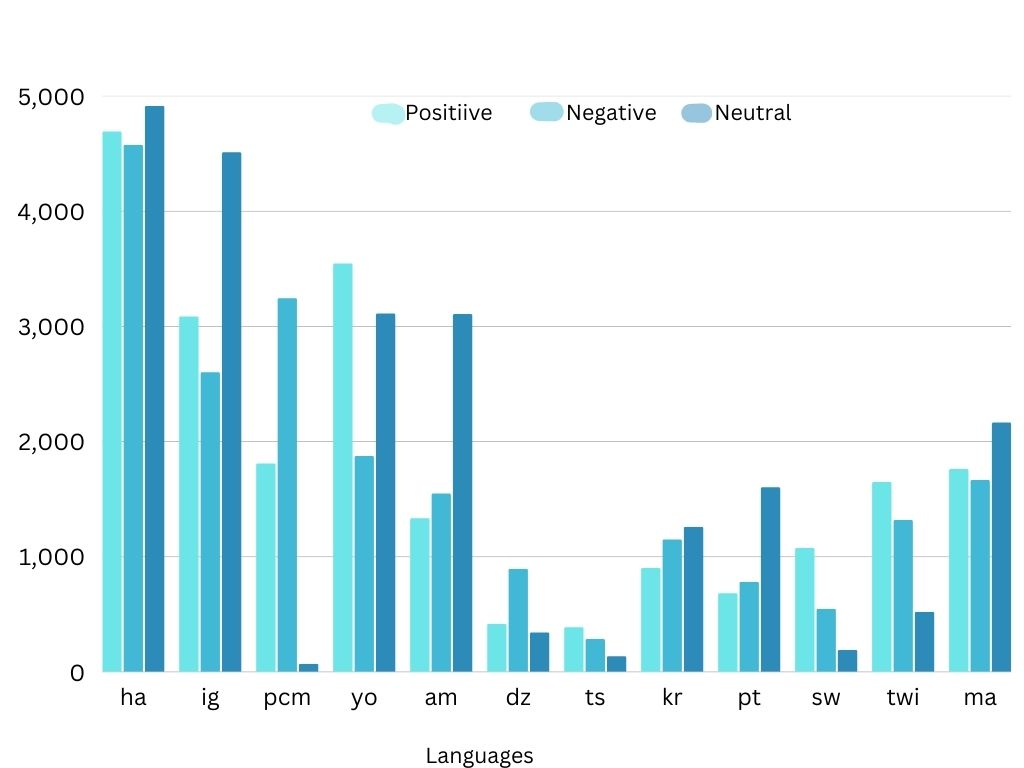}
    \vspace{0mm}
    
    \caption{\textbf{Distribution of Tweets Across Languages}. We show the graphical comparison of the various tweets across the 12 Languages for Monolingual Subtask A} 
    \label{fig:datasetdistro}
\end{figure*}

The AfriSenti dataset is a collection of Twitter datasets for sentiment analysis of African languages. The dataset used for the AfriSenti-SemEval 2023 shared task 12 consists of 14 languages: Hausa, Yoruba , Igbo,  Nigerian-Pidgin, Amharic, Algerian Arabic, Moroccan Arabic/Darija, Swahili, Kinyarwanda, Twi, Mozambican Portuguese, Setswana, isiZulu, Tigrinya, Xitsonga, and Oromo. The datasets are gold standard with multi-class labels (positive, negative, and neutral). Each tweet is annotated by three annotators following the annotation guidelines in \cite{mohammad2016practical} as shown in Table~\ref{tab:datasetdistro} and Figure~\ref{fig:datasetdistro}. Table~\ref{tab:datasetdistro} shows the distribution of the languages datasets,  sentiment labels, and sizes.

\begin{table*}[!t]
\centering
\begin{tabular}{ll|l|l|l|l|l|l}
\hline
\textbf{Subtask A: Monolingual} & \textbf{Pos} &\textbf{Pos\%} & \textbf{Neg} &\textbf{Neg\%} & \textbf{Neu} &\textbf{Neu \%}& \textbf{Total}\\
\hline

Amharic(am) &1332 &22.26 &1548 &25.87 &3104 &51.88 &05984 \\
Algerian Arabic(dz) &417 &25.26 &892 &54.03 &342 &20.72 &01651 \\
Hausa(ha) &4687 &33.08 &4573 &32.27 &4912 &34.66 &14172 \\
Igbo(ig) &3084 &30.26 &2600 &25.52 &4508 & 44.24 &10192 \\
Kinyarwanda(kr) &899 &27.23 &1146 &34.71 &1257 &38.07 &03302 \\
Darija(ma) &1758 &31.49 &1664 &29.81 &2161 &38.71 &05584 \\
Naija(pcm) &1808 &35.31 &3241 &63.29 &72 &1.41 &05121 \\
Mozambiqan Portuguese(pt) &681 &22.24 &782 &25.54 &1600 &52.24 &03063 \\
Swahili(sw) &1072 &59.23 &547 &30.23 &191 &10.56 &01810 \\

Xitsonga(ts) &384 &47.77 &284 &35.33 &136 &16.92 &00804 \\

Twi(twi) &1644 &47.23 &1315 &37.78 &522 &15.00 &03481 \\

Yorùbá(yo) &3542 &41.57 &1872 &21.97 &3108 &36.48 &08522 \\
\hline

\textbf{Subtask B: MultiLingual}           \\   
\hline 
Multilingual	&20783 &32.63 &20108 &31.57 &22794 &35.79	&63685

\\\hline

\textbf{Subtask C: Zero Shot}           \\   
\hline 
Tigrinya (Ti) & \multicolumn{6}{c|}{-}	&398\\
Oromo (Or) & \multicolumn{6}{c|}{-}	&396
\\ \hline
\end{tabular}

\caption{\textbf{Distribution of tweets across Languages}. Showing the balance distribution of various tweets across the 14 languages for Monolingual Subtask A, Multilingual Subtask B and Zero Shot Subtask C} 
\label{tab:datasetdistro}
\end{table*}

\section{Proposed Approach}
In this section, we describe our proposed approach for the SemEval shared task i.e leveraging low-resource tweet data for sentiment analysis of African languages. Our goal is to identify the sentiment classification of low-resource African languages using AfriBERTa-large, BERT-base-multilingual-cased and Afro-xlmr-large models. We have also shared the code on github \footnote{\url{https://github.com/ahmadmwali/SemEval-AfriSenti}}.

\subsection{Models Description}

\subsubsection{Afro-xlmr-large}
Afro-xlmr-large \cite{r13_alabi2022adapting} was created by Masked Language Modelling (MLM) adaptation of XLM-R-large model on 17 African languages (Afrikaans, Amharic, Hausa, Igbo, Malagasy, Chichewa, Oromo, Nigerian Pidgin, Kinyarwanda, Kirundi, Shona, Somali, Sesotho, Swahili, isiXhosa, Yoruba and isiZulu), covering the major African language families and 3 high resource languages (Arabic, French, and English). 
Afro-xlmr-large is used to facilitate sentiment classification of tweets in the low-resource African languages included in the AfriSenti shared task 12. The motivation behind using Afro-xlmr-large is that it is a multilingual model that is adapted to cover a wide range of African languages, including low-resource ones.
The multilingual model was created to help overcome the challenge of insufficient data in low-resource languages. However, one potential weakness of Afro-xlmr-large is that it may not perform as well in sentiment analysis tasks for languages that are not included in the model and the model usually requires large amounts of computational resources which is a limiting factor.

\subsubsection{BERT} 
Bert language model (Bidirectional Encoder Representations from Transformers) was introduced by
researchers from Google in 2018 and has become a popular and effective approach in natural
language processing tasks such as sentiment analysis, named entity recognition and question-
answering \cite{r7_devlin2018bert}. Bert has achieved state-of-the-art performance in various
benchmarks, and its pre-training on large corpora of text has shown to be effective in capturing
contextual information of words.
However, one limitation of Bert is its computational cost and memory requirements. The model
architecture is complex, and training on large amounts of data can be computationally expensive
and time-consuming. Another limitation is its vulnerability to adversarial attacks, where small
perturbations in the input text can lead to significant changes in the output prediction.
Bert has shown to be effective in capturing the nuances and complexities of language, especially in
languages with rich morphology and syntax. \cite{inoue2021interplay} used Bert-based models trained on
Arabic text data and achieved state-of-the-art performance on Arabic sentiment analysis
benchmarks. Similarly, AfroXLMR-large \cite{r13_alabi2022adapting} and MasakhaNER-1.0/2.0 \cite{10.1162/tacl_a_00416} used Bert-based models for named entity recognition and achieved high accuracy on African
language datasets. We experiment with multiple pre-trained BERT based models to competitively select the best across the datasets:

\paragraph{Bert-base-arabic-camelbert-da-sentiment}: Bert-base-arabic-camelbert-da-sentiment \cite{inoue2021interplay} is a collection of BERT models pre-trained on Arabic texts with different sizes and variants. Arabic Sentiment Tweet Dataset (ASTD) \cite{nabil2015astd}, an Arabic Speech-Act and Sentiment Corpus of Tweets (ArSAS) \cite{elmadany2018arsas} and SemEval datasets are used for fine-tuning the model.

\paragraph{Afroxlmr-large-ner-masakhaner-1.0-2.0}: masakhane/afroxlmr-large-ner-masakhaner-1.0-2.0 \cite{10.1162/tacl_a_00416} is a Named Entity Recognition (NER) model for 21 African languages. Specifically, this model is a Davlan/afro-xlmr-large model that was fine-tuned on an aggregation of African language datasets obtained from two versions of MasakhaNER dataset i.e. MasakhaNER 1.0 and MasakhaNER 2.0. One major advantage of using this model is that it has been trained on a wide range of African
languages and has been fine-tuned on datasets specific to those languages. This means that it is well-
suited for analysing sentiment in African languages, which can be challenging for other models as Bert is
trained on Arabic language and Afroxmlr-large does not cover 5 other languages.

\paragraph{Multilingual Bert (mBERT)}:  mBERT (Devlin et al., 2018) is a multilingual version of BERT pretrained on top 104 large language dataset from Wikipedia. It uses Masked Language Modeling (MLM).

\paragraph{AfriBERTa-Large}: The AfriBERTa-Large \cite{r9_ogueji2021small} is trained on mBERT using 11 African languages namely, Afaan Oromoo (also called Oromo), Amharic, Gahuza (a mixed language containing Kinyarwanda and Kirundi), Hausa, Igbo, Nigerian Pidgin, Somali, Swahili, Tigrinya and Yorùbá. It outperformed mBERT and XLM-R on several languages and is very competitive overall.

Table~\ref{tab:results} reports the results of the Large Language Models (LLMs) using Weighted F1 metric. Extensive experiments were carried out before selecting the best across the datasets and subtasks. Afro-XLMR-Large Performed best in most of the languages. Also Notably is Bert-base-arabic-camelbert-da-sentiment performance in arabic based datasets.

\section{Experiment and Evaluation}
This section describes the experimental and evaluation settings for our proposed approach for the SemEval- 2023 Task 12. 

\subsection{Experimental Settings}
The 3 subtasks use different training and testing size. We fine-tune the models with the provided training data and tune several hyper-parameters to obtain the optimal performances as shown in Table~\ref{tab:experiment}. Subtask C uses the same hyper parameters as Subtask A.

\begin{table}
\centering
\begin{tabular}{lll}
\hline
\textbf{Hyper Parameters} & \textbf{Subtask A} & \textbf{Subtask B}\\
\hline
Max-Length  & 128 & 150\\
Batch Size  & 16 & 32\\
Epoch  & 5 & 10\\
Optimizers  & AdamW & AdamW\\
Learning Rate  & 1e-5 & 2e-5\\
\hline 
\end{tabular}

\caption{Subtasks Hyper-Parameter Set}
\label{tab:experiment}
\end{table}

All subtasks will be evaluated using standard evaluation metric of weighted F1.

\section{Results and Discussion}

Table~\ref{tab:results} represents the performance of the 14 African languages evaluated by the weighted F1 metric. 

\begin{table*}[!t]
\centering
\begin{tabular}{r|p{0.6in}p{0.6in}p{0.5in}p{0.5in}p{0.5in}p{0.5in}}
\hline
\textbf{DATASETS} & \multicolumn{6}{c}{\textbf{PERFORMANCE ON WEIGHTED F1 METRIC}}\\
\textbf{Subtask A: Monolingual} & \textbf{AfriBERTa Large} &\textbf{Afroxlmr-Large} & \textbf{**arabic-camelbert} & \textbf{BERT} & \textbf{mBERT} &\textbf{Average}\\
\hline
am  &50.80	&57.30	&50.10	&\textbf{70.00}	&54.30	&56.50 \\
dz 	&54.60	&64.50	&\textbf{65.10}	&64.00	&54.70 &60.58\\
ha 	&79.50	&\textbf{81.00}	&69.20	&66.00	&76.80 &\textbf{74.50}\\
ig 	&\textbf{77.00}	&73.30	&68.00	&65.00	&73.30 &71.32\\
kr 	&50.90	&\textbf{70.60}	&51.30	&34.00	&65.70 &56.50\\
ma	&58.20	&\textbf{58.50}	&\textbf{58.50}	&45.00	&\textbf{58.50} &55.74\\
pcm	&64.20	&\textbf{68.50}	&63.70	&45.00	&66.60 &63.60\\
pt	&66.70	&\textbf{68.50}	&51.40	&68.00	&64.40 &63.80\\
sw	&\textbf{63.20}	&58.10	&56.90	&37.00	&57.30 &54.50\\
ts	&42.90	&\textbf{50.30}	&40.40	&37.00	&44.00 &44.92\\
twi	&\textbf{64.10}	&48.00	&55.70	&44.00	&59.30 &54.22\\
yo 	&62.90	&\textbf{71.90}	&60.15	&60.00	&67.80 &64.55\\
\hline

\textbf{Subtask B: Multilingual}           & \\   
\hline 
Multilingual	&69.30	&\textbf{*69.50}	&62.00	&66.00	&60.82	&\textbf{64.53}\\

\hline

\textbf{Subtask C: Zero Shot}           & \\   
\hline 
Ti &\textbf{62.50}	&55.00	&58.30	&56.70	&56.70	&\textbf{57.84}\\
Or &41.20	&\textbf{46.20}	&34.50	&39.50	&32.80	&38.84\\

\hline 
\textbf{Average} &60.53	&\textbf{62.26}	&56.35	&53.15	&59.53	 &58.40\\

\hline 
\end{tabular}

\caption{\textbf{Sentiment Classification Performance for the Three Subtasks.} Subtask A - Monolingual, Subtask B - Multilingual and Subtask C - Zero Shot. *Afroxlmr-large-ner-masakhaner-1.0-2.0 version of Afroxlmr is used for the Multilingual Dataset. **Bert-base-arabic-camelbert-da-sentiment}
\label{tab:results}
\end{table*}
For the monolingual subtask, the best dataset performance across the Pre-trained models is the Hausa language with an Average of 74.50\%, followed by Igbo and Yoruba with 71.32\% and 64.55.\% respectively. In terms of model performance, each of the pre-trained model performed best in at least one of the monolingual dataset. However, Afro-xlmr-Large ranked highest in best performances in 7 monolingual datasets, followed by AfriBERTa-Large and Arabic-Camelbert with 3 and 2 best performances respectively. BERT and mBERT with least best performances of 1 each. It is noteworthy that Arabic-Camelbert pre-trained model performed best in Darija and Moroccan Arabic due to the fact that it was originally trained in Arabic datasets. Likewise for Afro-xlmr-Large and AfriBERTa-Large which were largely trained in several Afro-centric datasets.

For multilingual subtask, Afro-xlmr-Large also performed best. This is much expected since the dataset is composed of all the 12 languages in the Monolingual subtask. The impressive 69.50\% performance is also attributed to the large volume of dataset. AfriBERTa-Large performance was also impressive with 69.30\% just 0.20\% difference. This performance shows that there’s a potential for building a cross-lingual model for a more advanced NLP system. 

For zero shot subtask, with two datsets, AfriBERTa-Large and Afro-xlmr-Large share the top spots with each emerging best in 1 of the tracks. AfriBERTa-Large dominated the Tigrinya zero shot track with an impressive 62.50\% performance. While Afro-xlmr-Large still claim another excellent performance on Oromo dataset track with 46.20\% as the best score.

Lastly, to average the performance across all the three subtasks, Afro-xlmr-Large came top with overall top score of 62.26\%. It was trailed behind by AfriBERTa-Large with 60.53\%. mBERT, Arabic-CamelBERT and BERT coming distant third with 59.53.00\%, fourth with 56.35\% and fifth with 53.12\% respectively.

\subsection{Ablation Study}

We further perform an ablation study to attribute the reason for the performance across the datasets and subtasks. 
For Monolingual, the Xitsonga language achieved the least best performance with a weighted F1 of 50.3\%. While Hausa achieved the best of the bests performance with 81.00\%. 

As shown in Table~\ref{tab:datasetdistro}, the Xitsonga language has a smaller volume of data of just 804 training size and thus the lower performance. While Hausa has 14172 and achieved a superior performance across all the five pre-trained models.  Therefore, we are of the opinion that there is a correlation between performance and volume of data for a given language. There is better performance when implemented in languages with larger datasets than in smaller dataset.


\section{Conclusion and Future Work}
In this paper, we presented our system description for the SemEval shared task on sentiment analysis for low-resource African languages using Twitter dataset. The task consists of three sub-tasks: Monolingual, Multilingual, and Zero-Shot.
Several pretrained LLMs were used for fine-tuning. Afro-xlmr-large performed relatively best across the three subtasks coming top in 9 out of 15. AfriBERTa came top in 4, while Bert-base-arabic-camelbert-da-sentiment performed best in arabic datasets of Darija and Algerian Arabic. BERT and mBERT also manage to came top in 1 task each. Experimental results demonstrated that Nigerian languages: Hausa, Igbo and Yoruba achieved better performance compared to other languages due to the higher volume of data present in the languages. Our results indicate
that deep learning are effective in sentiment classification in the African language given the right data,
model and training. For future work, to incorporate other data sources such as moview reviews or news articles for generalizability. We also recommend fine tuning for specific individual languages to incorporate linguistic features specific to each language, such as idioms or colloquialisms.

\section*{Acknowledgements}

We would like to acknowledge the support of HausaNLP Management for providing us with the Google Colab GPU Premium Version.

\bibliography{anthology,custom}
\bibliographystyle{acl_natbib}

\appendix




\end{document}